\documentclass[12pt]{article}

\usepackage[utf8]{inputenc}
\usepackage[T1]{fontenc}
\usepackage{times}
\usepackage[margin=1in]{geometry}
\usepackage{setspace}
\usepackage{natbib}
\usepackage{graphicx}
\usepackage{booktabs}
\usepackage{multirow}
\usepackage{amsmath}
\usepackage{hyperref}
\usepackage{xcolor}
\usepackage{float}
\usepackage{tipa}

\title{Phonological Fossils: Machine Learning Detection of Non-Mainstream Vocabulary in Sulawesi Basic Lexicon}

\author{Mukhlis Amien \and Go Frendi Gunawan}

\date{}

\begin{document}

\maketitle

\begin{abstract}
Basic vocabulary in many Sulawesi Austronesian languages includes forms that resist reconstruction to any proto-form and show phonological patterns inconsistent with inherited roots---but whether this non-conforming vocabulary represents pre-Austronesian substrate or independent innovation has not been tested computationally.
We address this question by combining a rule-based cognate subtraction method with a machine learning classifier trained exclusively on phonological features.
Using 1,357 lexical forms from six Sulawesi languages in the Austronesian Basic Vocabulary Database, we first identify 438 candidate substrate forms (26.5\% of the corpus) through cognate subtraction and Proto-Austronesian cross-checking.
An XGBoost classifier trained on 26 phonological and distributional features---excluding all cognacy data---distinguishes inherited Austronesian vocabulary from non-mainstream forms with AUC = 0.763 ($\pm$ 0.007), revealing a ``phonological fingerprint'': longer forms, more consonant clusters, higher rates of glottal stops, and fewer canonical Austronesian prefixes.
This fingerprint is robust to approximate IPA conversion (AUC change $<$ 0.01), confirming that the model detects phonological patterns rather than orthographic artifacts.
Cross-method consensus analysis (Cohen's $\kappa$ = 0.61) identifies 266 high-confidence non-mainstream candidates confirmed by both methods.
However, phonological clustering of these consensus candidates yields no coherent word families (silhouette = 0.114; cross-linguistic cognate test $p$ = 0.569), providing no evidence that they derive from a single pre-Austronesian language layer.
Application to 16 additional languages confirms geographic patterning: Sulawesi languages show significantly higher predicted non-mainstream rates (mean $P_\mathrm{sub}$ = 0.606) than Western Indonesian languages (0.393).
This study demonstrates that phonological machine learning can complement traditional comparative methods in detecting non-mainstream lexical layers, while cautioning against interpreting phonological non-conformity as evidence for a shared substrate language.
\end{abstract}


\section{Introduction}

The Austronesian language family, comprising over 1,200 languages spoken from Madagascar to Easter Island, represents one of the most extensively documented and well-understood language families in the world \citep{blust2013}.
Yet across many Austronesian languages, a substantial proportion of basic vocabulary resists reconstruction to any proto-form: these forms lack plausible cognates, deviate from canonical disyllabic root structure, and show phonological patterns inconsistent with established sound correspondences.
In Island Southeast Asia---particularly Sulawesi, with its extraordinary linguistic diversity spanning at least ten primary subgroups \citep{blust2009}---such phonologically non-conforming vocabulary has traditionally been interpreted as evidence of pre-Austronesian substrate languages displaced by Austronesian expansion \citep{donohue2010, adelaar2005}.
But this interpretation assumes that phonological non-conformity implies shared non-Austronesian ancestry---an assumption that has rarely been tested systematically.

The traditional approach to substrate detection relies on the comparative method: systematic identification of forms that resist reconstruction to any proto-language, that lack plausible cognates across related languages, and that show phonological patterns inconsistent with established sound correspondences \citep{blust2009, thomason1988}.
While powerful, this approach requires deep specialist knowledge of each language under investigation and scales poorly across large datasets.
As the field increasingly adopts computational tools for phylogenetic analysis \citep{gray2009, list2018}, automated cognate detection \citep{list2012}, global-scale lexical inference \citep{jaeger2018}, and cross-linguistic comparison \citep{forkel2018}, there is an opportunity to apply machine learning methods to the substrate detection problem.

Sulawesi provides an ideal testing ground for such an approach.
The island's languages span multiple primary-level Austronesian subgroups, including Celebic, South Sulawesi, and Muna--Buton, yet share a deep geographic co-existence \citep{himmelmann2005}.
The island's complex geological and demographic history---shaped by volcanism, maritime networks, and waves of migration---creates conditions under which substrate retention is plausible \citep{bellwood1995}.
Moreover, the Austronesian Basic Vocabulary Database \citep[ABVD;][]{greenhill2008} provides standardized cognate-coded Swadesh lists for numerous Sulawesi languages, offering a computationally tractable dataset.

This study makes two contributions, the second more consequential than the first.
First, we introduce a machine learning methodology for detecting phonological non-conformity that addresses the circularity inherent in using cognacy data to identify non-cognate forms.
Our two-model design separates a ``circular'' baseline (Model~A, including cognacy features) from the genuine experiment (Model~B, using only phonological features), establishing that phonological properties alone carry detectable signal about a form's status as inherited or non-mainstream vocabulary.
Second---and more importantly---we test the assumption that underlies substrate interpretations: that non-conforming vocabulary across related languages derives from a common non-Austronesian source.
Phonological clustering of the detected candidates finds no support for this assumption ($p$ = 0.569): the non-mainstream vocabulary shows no evidence of shared etymological descent, consistent with parallel independent innovations rather than remnants of a single pre-Austronesian language layer.
This negative result constrains the interpretation of both computational and traditional substrate detection: phonological non-conformity is not equivalent to non-Austronesian origin.

\section{Data and Methods}

\subsection{Data}
\label{sec:data}

We draw on the Austronesian Basic Vocabulary Database \citep{greenhill2008}, accessed in Cross-Linguistic Data Format \citep[CLDF;][]{forkel2018}.
Our primary dataset comprises 1,357 lexical forms from six Sulawesi languages: Muna ($n$ = 219), Bugis ($n$ = 242), Makassar ($n$ = 217), Wolio ($n$ = 254), Toraja-Sa'dan ($n$ = 216), and Tolaki ($n$ = 209).
These languages represent three subgroups---Muna--Buton (Muna, Wolio), South Sulawesi (Bugis, Makassar), Celebic (Toraja-Sa'dan), and Southeast Sulawesi (Tolaki)---providing typological breadth within the Sulawesi Austronesian context.

Each form covers one of 210 Swadesh concepts, with associated cognate set codes from the ABVD.
Following \citet{swadesh1955}, we distinguish a core Swadesh-100 subset (the most resistant to borrowing) from the extended Swadesh-210 list.
Forms were cleaned by removing bracketed annotations, leading/trailing punctuation, and whitespace artifacts.

Labels were assigned through the rule-based subtraction method described in Section~\ref{sec:rule_based}: forms retaining ABVD cognacy codes (or rescued through Proto-Austronesian cross-checking) were labeled ``Austronesian'' ($n$ = 919, 67.7\%); residual forms without cognacy attribution were labeled ``candidate substrate'' ($n$ = 438, 32.3\%).
We treat this as a Positive-Unlabeled (PU) learning problem: the ``Austronesian'' label is reliable (positive), while ``candidate substrate'' contains both genuine non-Austronesian forms and Austronesian forms with missing cognacy data (unlabeled positives).

\subsection{Rule-Based Baseline (E022)}
\label{sec:rule_based}

Our rule-based substrate detection follows a subtractive logic.
For each language, we began with the full ABVD Swadesh list and removed three layers:

\begin{enumerate}
    \item \textbf{ABVD cognates:} Forms assigned to any cognate set in the ABVD were classified as Austronesian.
    \item \textbf{Known loanwords:} Forms matching established Sanskrit, Arabic, or Malay trade loanword lists were excluded from the residual.
    \item \textbf{Proto-Austronesian cross-check:} 15 known Proto-Austronesian reconstructions that the ABVD had missed cognacy codes for were identified through manual comparison with \citet{blust2013}, rescuing 75 false positive residuals across the six languages.
\end{enumerate}

After this three-stage subtraction, the average residual rate was 26.5\% (Table~\ref{tab:residuals}), ranging from 11.9\% (Muna) to 54.5\% (Tolaki).
Eight concepts appeared as residuals in five or more of the six languages (Tier~1 candidates): ``if'', ``to bite'', ``to tie up'', ``to cut/hack'', ``grass'', ``to throw'', ``they'', and ``big''.
These cross-linguistically persistent residuals represent the strongest candidates for pre-Austronesian substrate or shared innovation.

Tolaki's elevated residual rate (54.5\%) reflects its low ABVD cognacy coverage (36\% of forms assigned to cognate sets), likely an artifact of under-documentation rather than genuine substrate enrichment.

\subsection{ML Classification (E027)}
\label{sec:ml}

\subsubsection{Two-model design}

To address the circularity of using cognacy-derived labels to train a classifier, we employed a two-model design:

\begin{itemize}
    \item \textbf{Model~A (full features):} Included 31 features encompassing distributional properties (cognate set size, number of cognate sets per form, concept-level residual rate, cross-language residual count) alongside phonological features. This model was expected to achieve near-perfect performance, since the distributional features encode the labeling criterion. Model~A serves as a ranking tool, not as a scientific claim.
    \item \textbf{Model~B (phonological-only):} Used 27 features, excluding all distributional/cognacy-derived features. If this model achieves above-chance performance, it demonstrates that phonological properties alone distinguish substrate candidates from inherited vocabulary---a non-circular finding.
\end{itemize}

\subsubsection{Feature engineering}

Model~B's 27 features fall into four categories:

\begin{enumerate}
    \item \textbf{Phonological features (10):} Form length (syllable count, approximated as the number of vowel nuclei; validation confirms equivalence with character count, Section~\ref{sec:results_ipa}), vowel count, vowel ratio, whether the form ends in a vowel, initial character class (one-hot: /m/, /a/, /b/, /t/, /k/, /p/, /s/, other), presence of glottal stop markers (\textipa{P} or \texttt{'}), presence of nasal clusters (\textit{ng}, \textit{mb}, \textit{nd}, \textit{nj}, etc.), presence of reduplication (detected via hyphenation or repeated bigram/trigram patterns---a surface-level approximation that does not capture all morphological reduplication types), consonant cluster count (CC+ sequences in surface orthography; note that clusters at morpheme boundaries, e.g.\ from infixation, are not distinguished from root-internal clusters), and presence of Austronesian-like prefixes (\textit{ma-}, \textit{me-}, \textit{mo-}, \textit{pa-}, \textit{ka-}, etc.; infixes and suffixes are not detected).

    \item \textbf{Initial-character one-hot encoding (8):} Seven frequent onset classes (/m/, /a/, /b/, /t/, /k/, /p/, /s/) plus ``other'', capturing onset frequency distributions that may differ between inherited and substrate vocabulary.

    \item \textbf{Semantic features (2):} Binary core vocabulary membership (Swadesh-100 vs.\ extended Swadesh-210), and semantic domain (one-hot: \textsc{action}, \textsc{body}, \textsc{nature}, \textsc{quality}, \textsc{number}, \textsc{grammar}, \textsc{other}).

    \item \textbf{Language control features (2):} Language identity (integer-encoded) and language-level cognacy coverage rate (proportion of forms with ABVD cognate codes), controlling for between-language variation in documentation depth.
\end{enumerate}

All features were computed from surface orthographic forms in the ABVD without morphological decomposition, not from phonemic or IPA transcriptions.
This introduces known limitations---orthographic bias and inability to distinguish root-internal from morphological consonant clusters (Section~\ref{sec:limitations})---but ensures full reproducibility using publicly available data.
Robustness tests (Section~\ref{sec:results_ipa}) confirm that approximate IPA conversion, syllable-based length measurement, and even complete removal of the length feature produce equivalent model performance.

\subsubsection{Classifiers}

We trained three classifiers to assess robustness:

\begin{itemize}
    \item \textbf{XGBoost} \citep{chen2016}: 300 estimators, max depth 4, learning rate 0.05, with scale\_pos\_weight adjusted for class imbalance.
    \item \textbf{Random Forest} \citep{breiman2001}: 500 estimators, minimum 5 samples per leaf, class-balanced sample weights.
    \item \textbf{Logistic Regression}: L2 regularization, class-balanced weights, serving as a linear baseline.
\end{itemize}

All classifiers were implemented using scikit-learn \citep{pedregosa2011} and XGBoost \citep{chen2016}.

\subsubsection{Validation}

Two validation protocols were employed:

\begin{enumerate}
    \item \textbf{Stratified 5-fold cross-validation}, repeated across 10 random seeds, yielding 50 train--test splits per classifier. We report mean AUC, F1, and accuracy with standard deviations.
    \item \textbf{Leave-One-Language-Out (LOLO):} Each of the six languages is held out as the test set in turn, with the remaining five used for training. This tests whether the phonological fingerprint generalizes across languages rather than memorizing language-specific patterns.
\end{enumerate}

\subsubsection{Interpretability}

We used SHAP (SHapley Additive exPlanations; \citealp{lundberg2017}) to decompose Model~B predictions into per-feature contributions, identifying which phonological properties most strongly drive substrate classification.

\subsection{Cross-Method Consensus (E028)}
\label{sec:consensus}

To identify high-confidence substrate candidates, we cross-tabulated the rule-based (E022) and ML (E027 Model~B) predictions into a four-quadrant framework:

\begin{itemize}
    \item \textbf{CS (Consensus Substrate):} E022 residual \textsc{and} ML $P \geq 0.5$. High-confidence substrate.
    \item \textbf{CA (Consensus Austronesian):} E022 cognate \textsc{and} ML $P < 0.5$. High-confidence Austronesian.
    \item \textbf{RO (Rule-Only):} E022 residual \textsc{but} ML $P < 0.5$. Probable E022 false positives.
    \item \textbf{MO (ML-Only):} E022 cognate \textsc{but} ML $P \geq 0.5$. Potential missed substrates.
\end{itemize}

Agreement was quantified using Cohen's $\kappa$ \citep{cohen1960}, with Pearson and Spearman correlations between $P_\mathrm{substrate}$ and the E022 binary label.
Following \citet{landis1977}, we interpret $\kappa$ values in the ranges: $< 0.20$ (slight), 0.21--0.40 (fair), 0.41--0.60 (moderate), 0.61--0.80 (substantial), 0.81--1.00 (almost perfect).

\subsection{Phonological Clustering (E029)}
\label{sec:clustering}

If the consensus substrate candidates originate from a single pre-Austronesian language, forms denoting the same concept across different Sulawesi languages should be phonologically more similar to each other than to random vocabulary.
We tested this prediction as follows.

\subsubsection{Distance metric}

Pairwise normalized Levenshtein edit distance \citep{levenshtein1966} was computed over orthographic forms, yielding values from 0 (identical) to 1 (maximally different).

\subsubsection{Clustering algorithms}

Two complementary approaches were applied to the 266 consensus substrates:

\begin{itemize}
    \item \textbf{Ward's hierarchical agglomerative clustering} \citep{ward1963}: silhouette scores were computed for $k = 5$ to $k = 30$ to identify the optimal number of clusters.
    \item \textbf{DBSCAN} \citep{ester1996}: density-based clustering with grid search over $\varepsilon \in \{0.3, 0.35, 0.4, 0.45, 0.5\}$ and $\mathrm{min\_samples} \in \{3, 4, 5\}$.
\end{itemize}

\subsubsection{Cross-linguistic cognate test}

For 20 concepts appearing as consensus substrates in three or more languages, we computed the mean pairwise Levenshtein distance across languages and compared it to a null distribution of 1,000 random concept forms from the same languages.
A one-tailed test assessed whether substrate forms are more similar (lower distance) than random forms, as expected under the shared-substrate hypothesis.

\section{Results}

\subsection{Rule-Based Residuals}
\label{sec:results_e022}

Table~\ref{tab:residuals} presents per-language residual rates after cognate subtraction with Proto-Austronesian cross-checking.

\begin{table}[H]
\centering
\caption{Per-language residual rates after rule-based cognate subtraction (E022).}
\label{tab:residuals}
\begin{tabular}{lrrrrr}
\toprule
Language & Total & Has Cognacy & Residual & \% Residual & Coverage \\
\midrule
Muna         & 219 & 185 (84\%) &  26 & 11.9\% & 84\% \\
Bugis        & 242 & 180 (74\%) &  49 & 20.2\% & 74\% \\
Toraja-Sa'dan & 216 & 171 (79\%) &  32 & 14.8\% & 79\% \\
Wolio        & 254 & 171 (67\%) &  68 & 26.8\% & 67\% \\
Makassar     & 217 & 137 (63\%) &  67 & 30.9\% & 63\% \\
Tolaki       & 209 &  75 (36\%) & 114 & 54.5\% & 36\% \\
\midrule
\textbf{Mean} & --- & --- & --- & \textbf{26.5\%} & --- \\
\bottomrule
\end{tabular}
\end{table}

The 26.5\% average residual includes 75 forms rescued from false-positive status through Proto-Austronesian cross-checking.
Eight concepts persist as residuals in 5--6 of the six languages (``if'', ``to bite'', ``to tie up'', ``to cut/hack'', ``grass'', ``to throw'', ``they'', ``big''), representing the most robust substrate candidates.

\subsection{ML Classification}
\label{sec:results_ml}

\subsubsection{Cross-validation}

Table~\ref{tab:cv_results} presents stratified 5-fold cross-validation results across 10 random seeds for both models.

\begin{table}[H]
\centering
\caption{Stratified 5-fold CV results ($\times$ 10 seeds). Model~A includes cognacy features (circular baseline); Model~B uses only phonological features (scientific claim).}
\label{tab:cv_results}
\begin{tabular}{llccc}
\toprule
Model & Classifier & AUC & F1 & Accuracy \\
\midrule
\multirow{3}{*}{A (Full)} & XGBoost & 1.000 $\pm$ 0.001 & 1.000 $\pm$ 0.001 & 1.000 $\pm$ 0.001 \\
 & Random Forest & 1.000 $\pm$ 0.000 & 1.000 $\pm$ 0.000 & 1.000 $\pm$ 0.001 \\
 & Logistic Reg. & 1.000 $\pm$ 0.000 & 1.000 $\pm$ 0.000 & 1.000 $\pm$ 0.000 \\
\midrule
\multirow{3}{*}{B (Phon.)} & \textbf{XGBoost} & \textbf{0.760 $\pm$ 0.007} & \textbf{0.822 $\pm$ 0.005} & \textbf{0.741 $\pm$ 0.008} \\
 & Random Forest & 0.762 $\pm$ 0.006 & 0.788 $\pm$ 0.006 & 0.717 $\pm$ 0.008 \\
 & Logistic Reg. & 0.747 $\pm$ 0.003 & 0.748 $\pm$ 0.005 & 0.683 $\pm$ 0.006 \\
\bottomrule
\end{tabular}
\end{table}

Model~A's near-perfect performance confirms that distributional features (cognate set size, cognate count) encode the label directly---this model is circular and scientifically uninformative, serving only as a ranking tool.
Model~B's AUC of 0.760 demonstrates that phonological properties alone carry moderate but reliable discriminative signal, independent of cognacy data.
The three classifiers yield similar AUCs (0.747--0.762), suggesting the signal is robust to algorithmic choice.

\subsubsection{Leave-One-Language-Out}

Table~\ref{tab:lolo} presents LOLO results for Model~B (XGBoost), testing cross-linguistic generalization.

\begin{table}[H]
\centering
\caption{Leave-One-Language-Out results for Model~B (XGBoost). Each language was held out as the test set in turn.}
\label{tab:lolo}
\begin{tabular}{lrrrrr}
\toprule
Held-out language & AUC & F1 & Acc. & $N_\mathrm{test}$ & $N_\mathrm{substr}$ \\
\midrule
Tolaki       & 0.806 & 0.530 & 0.364 & 209 & 134 \\
Makassar     & 0.747 & 0.781 & 0.659 & 217 &  80 \\
Bugis        & 0.727 & 0.799 & 0.707 & 242 &  62 \\
Wolio        & 0.697 & 0.798 & 0.669 & 254 &  83 \\
Toraja-Sa'dan & 0.696 & 0.787 & 0.690 & 216 &  45 \\
Muna         & 0.618 & 0.465 & 0.402 & 219 &  34 \\
\midrule
\textbf{Mean} & \textbf{0.715 $\pm$ 0.057} & --- & --- & --- & --- \\
\bottomrule
\end{tabular}
\end{table}

Five of six languages achieve AUC $\geq$ 0.65, indicating that the phonological fingerprint generalizes across Sulawesi languages.
Muna (AUC = 0.618) is the weakest, likely because its low residual rate (11.9\%) provides limited substrate signal for the held-out test.
Tolaki achieves the highest AUC (0.806) despite its inverted class balance (64\% substrate), consistent with its phonologically distinctive non-mainstream vocabulary.

\subsubsection{Sensitivity: Tolaki}

Because Tolaki contributes a disproportionate share of substrate candidates (134/438, 30.6\%), we assessed sensitivity by retraining Model~B without Tolaki.
The AUC decreased from 0.760 to 0.698 ($\Delta$ = $-$0.062) but remained above the 0.65 threshold, confirming that the phonological signal is not solely driven by Tolaki.

\subsection{Feature Ablation}
\label{sec:results_ablation}

Because \texttt{language\_cognacy\_coverage} is the highest-ranked SHAP feature and a language-level property correlated with the labeling process, we tested whether the phonological signal survives its removal.
We trained three model variants: (1)~the full 27-feature Model~B; (2)~an ablated model excluding \texttt{language\_cognacy\_coverage} (26 features); and (3)~a ``pure phonological'' model excluding both language-control features (25 features).

\begin{table}[H]
\centering
\caption{Feature ablation results (XGBoost). Removing the cognacy coverage confound \textit{improves} performance.}
\label{tab:ablation}
\begin{tabular}{lccr}
\toprule
Variant & CV AUC & LOLO mean AUC & LOLO $\geq$0.65 \\
\midrule
Full Model~B (27 features) & 0.760 $\pm$ 0.007 & 0.715 & 5/6 \\
\textbf{Ablated} ($-$coverage, 26 feat.) & \textbf{0.763 $\pm$ 0.007} & \textbf{0.722} & \textbf{6/6} \\
Pure phonological (25 feat.) & 0.727 $\pm$ 0.007 & 0.701 & 5/6 \\
\bottomrule
\end{tabular}
\end{table}

Removing \texttt{language\_cognacy\_coverage} slightly \textit{improves} both CV AUC ($+$0.003) and LOLO AUC ($+$0.007), with all six languages now exceeding the 0.65 threshold.
Muna, previously the weakest held-out language (AUC = 0.618), improves to 0.679 under ablation---the confound feature was actively harming cross-linguistic generalization for low-residual-rate languages.
The phonological fingerprint is therefore robust to removal of the cognacy coverage confound: the signal resides in form-level phonological properties, not in language-level documentation artifacts.

\subsection{IPA Robustness Test}
\label{sec:results_ipa}

Because all phonological features were computed from orthographic forms (Section~\ref{sec:limitations}), we tested whether approximate IPA conversion affects model performance.
Orthographic digraphs were converted to single IPA characters using conservative, language-specific mappings: \textit{ng} $\rightarrow$ /\textipa{N}/ and \textit{ny} $\rightarrow$ /\textipa{\textltailn}/ (all languages), plus \textit{gh} $\rightarrow$ /\textipa{G}/ and \textit{bh} $\rightarrow$ /\textipa{B}/ (Muna).
This affected 75/1,357 forms (5.5\%), predominantly in Muna (54 forms, 24.7\%).
Phonological features were recomputed on the IPA forms and Model~B was retrained.

IPA conversion produced negligible change in cross-validation performance (AUC 0.772 $\rightarrow$ 0.774, $\Delta$ = $+$0.002) and slight improvement in LOLO validation (mean AUC 0.724 $\rightarrow$ 0.733, $\Delta$ = $+$0.009).
All six LOLO languages maintained AUC $\geq$ 0.65 under IPA conversion.
Notably, Muna---the language most affected by digraph conversion---showed the largest LOLO improvement ($+$0.042), indicating that orthographic digraphs were adding noise rather than signal.
The phonological fingerprint is therefore robust to orthographic-to-IPA conversion: the model detects phonological patterns, not orthographic artifacts.

A related concern is that \texttt{form\_length} measures character count rather than the linguistically meaningful syllable count.
We tested this by replacing character count with a vowel-nuclei syllable counter: model performance was unchanged (CV AUC 0.768 $\rightarrow$ 0.769, $\Delta < 0.001$; LOLO 0.722 $\rightarrow$ 0.728, $\Delta$ = $+$0.006).
Indeed, removing the length feature entirely produces equivalent performance (CV AUC 0.769, LOLO 0.732), confirming that the phonological fingerprint does not depend on the length metric---it is carried by other features (consonant clusters, glottal stops, prefix patterns).

\subsection{SHAP Analysis}
\label{sec:results_shap}

\begin{figure}[H]
\centering
\includegraphics[width=0.85\textwidth]{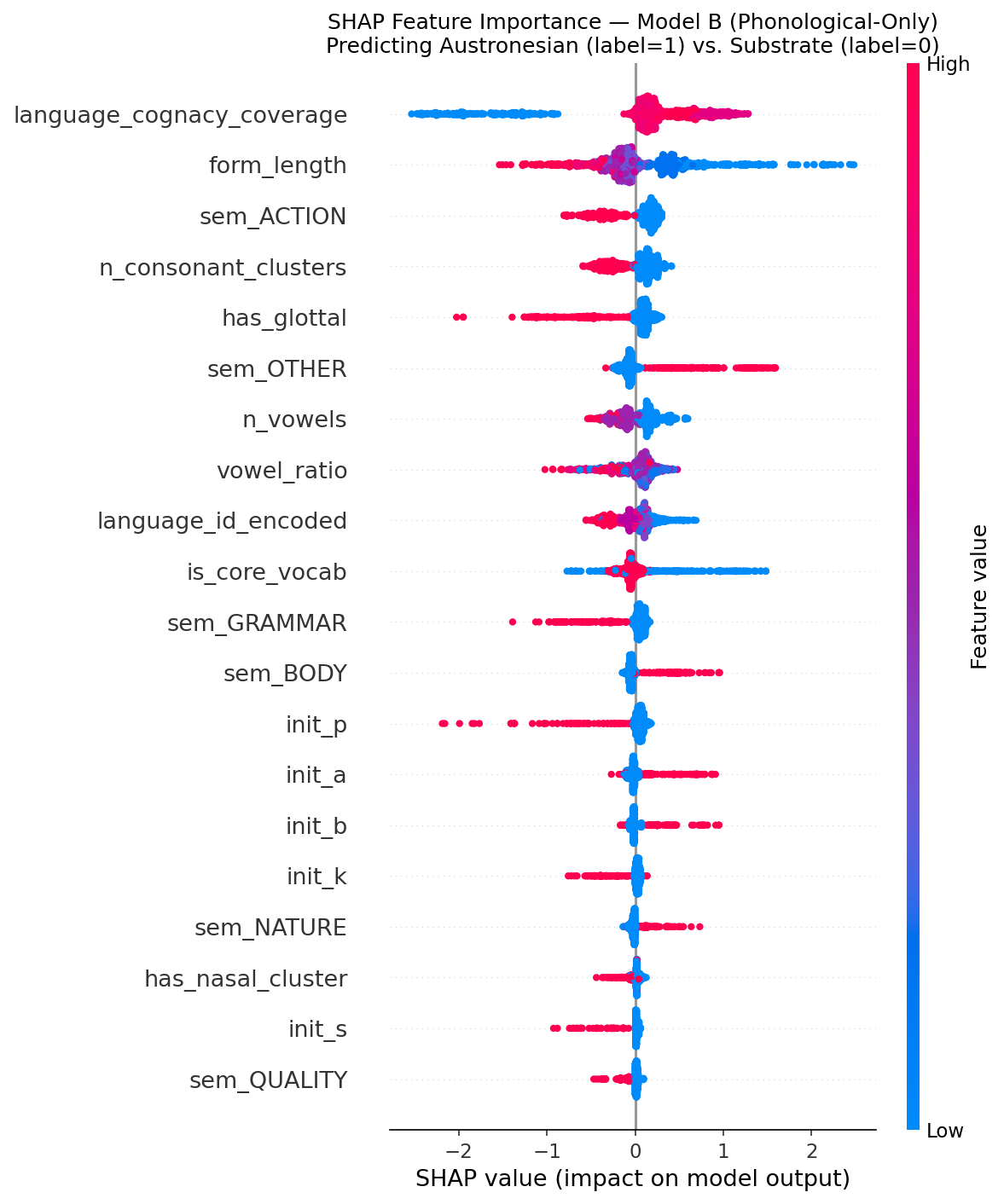}
\caption{SHAP beeswarm plot for Model~B (XGBoost). Each dot represents one form; color indicates feature value (red = high, blue = low). Positive SHAP values push toward substrate classification.}
\label{fig:shap}
\end{figure}

SHAP analysis of Model~B reveals the phonological features most strongly associated with substrate classification (Figure~\ref{fig:shap}).

The five most important features are:

\begin{enumerate}
    \item \textbf{Language cognacy coverage} (mean $|$SHAP$|$ = 0.559): Languages with lower ABVD coverage rates have higher predicted substrate probabilities. This is a language-level control variable that captures documentation depth rather than a phonological property per se.
    \item \textbf{Form length} (0.378): Non-mainstream candidates average 2.57 syllables compared to 2.29 for inherited forms ($+$0.28 syllables), consistent with the typological observation that Austronesian basic vocabulary tends toward disyllabic roots \citep{blust2009}. Character count and syllable count produce identical model performance ($\Delta$AUC $<$ 0.001), and removing the length feature entirely leaves performance unchanged (Section~\ref{sec:results_ipa}), indicating that length contributes modestly alongside other features.
    \item \textbf{Semantic domain: \textsc{action}} (0.230): Action verbs are over-represented among non-mainstream candidates. In the top 50 highest-confidence candidates, 46\% are action verbs, compared to 26\% quality adjectives and 16\% grammatical items. A caveat applies: action verbs in Sulawesi languages frequently undergo morphological derivation (infixation, suffixation) that may increase surface form complexity; the extent to which this semantic signal reflects substrate persistence versus morphological inflation cannot be resolved without morphological decomposition of the ABVD forms.
    \item \textbf{Consonant cluster count} (0.190): Non-mainstream candidates contain more CC+ sequences, deviating from the predominantly (C)V(C) canonical syllable structure of Austronesian languages. As with form length, some of these clusters may arise at morpheme boundaries (e.g.\ from infixation) rather than within roots; future work with morphologically parsed data could distinguish these sources.
    \item \textbf{Glottal stop presence} (0.188): Forms containing glottal stop markers (\textipa{P} or orthographic \texttt{'}) are more likely to be classified as substrate.
\end{enumerate}

We term the combination of these properties---longer forms, more consonant clusters, more glottal stops, fewer canonical prefixes, and action-verb semantics---the ``phonological fingerprint'' of non-mainstream vocabulary in Sulawesi basic lexicon.

\subsection{Cross-Method Consensus}
\label{sec:results_consensus}

Cross-tabulation of rule-based (E022) and ML (E027) predictions yields a four-quadrant distribution (Figure~\ref{fig:quadrant}):

\begin{figure}[H]
\centering
\includegraphics[width=0.85\textwidth]{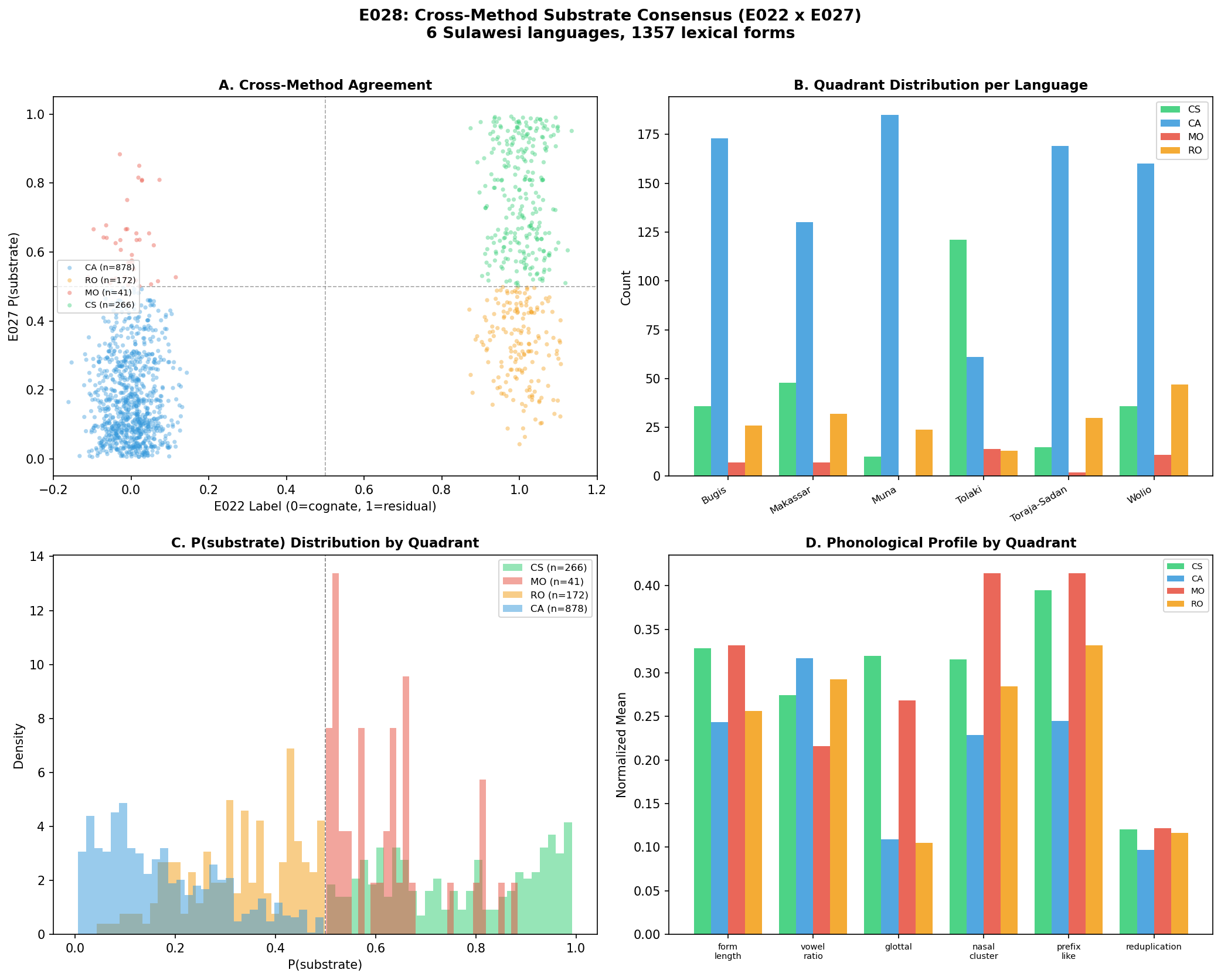}
\caption{Four-quadrant comparison of rule-based (E022) and ML (E027 Model~B) substrate predictions. CS = Consensus Substrate, CA = Consensus Austronesian, RO = Rule-Only, MO = ML-Only.}
\label{fig:quadrant}
\end{figure}

\noindent Cohen's $\kappa$ = 0.611, indicating substantial agreement \citep{landis1977}.
The quadrant distribution is as follows: 266 Consensus Substrate (CS, 19.6\%), 878 Consensus Austronesian (CA, 64.7\%), 172 Rule-Only (RO, 12.7\%), and 41 ML-Only (MO, 3.0\%).

The 266 CS forms represent 60.7\% of all E022 residuals, confirmed independently by ML phonological analysis.
Their distribution across languages mirrors the residual rate hierarchy: Tolaki (121), Makassar (48), Bugis (36), Wolio (36), Toraja-Sa'dan (15), Muna (10).

Semantic domain analysis of CS forms reveals a predominance of action verbs: \textsc{action} 117 (44.0\%), \textsc{grammar} 40 (15.0\%), \textsc{quality} 38 (14.3\%), \textsc{nature} 20 (7.5\%), \textsc{number} 20 (7.5\%), \textsc{body} 12 (4.5\%).

\subsubsection{Disagreement analysis}

The 172 RO forms (E022 residuals that the ML classifies as Austronesian-like) are characterized by shorter form length (mean 5.35 vs.\ 6.58 for CS), fewer glottal stops (10.5\% vs.\ 32.0\%), and fewer consonant clusters (0.35 vs.\ 0.56).
These are plausibly E022 false positives---short forms that lacked ABVD cognacy codes but are phonologically regular Austronesian vocabulary.

The 41 MO forms (E022 cognates that the ML flags as substrate-like) show the opposite pattern: longer forms (mean 6.63), more glottal stops (26.8\%), and substantially more consonant clusters (0.73 vs.\ 0.30 for CA).
These warrant investigation as potential missed substrates that received cognacy codes despite substrate-like phonology.

Five concepts appear as CS in four or more of the six languages: ``One Hundred'', ``Fifty'', ``Twenty'', ``to stand'', and ``to hit''.

\subsection{Clustering Results}
\label{sec:results_clustering}

Phonological clustering of the 266 consensus substrates reveals no coherent word families.
Ward's hierarchical clustering yields an optimal $k$ = 30 with a silhouette score of 0.114, indicating nearly random cluster assignment.
DBSCAN at its best configuration ($\varepsilon$ = 0.3, min\_samples = 3) finds only 4 tiny clusters containing 14 forms total, with 94.7\% classified as noise.

\subsubsection{Cross-linguistic cognate test}

\begin{figure}[H]
\centering
\includegraphics[width=0.75\textwidth]{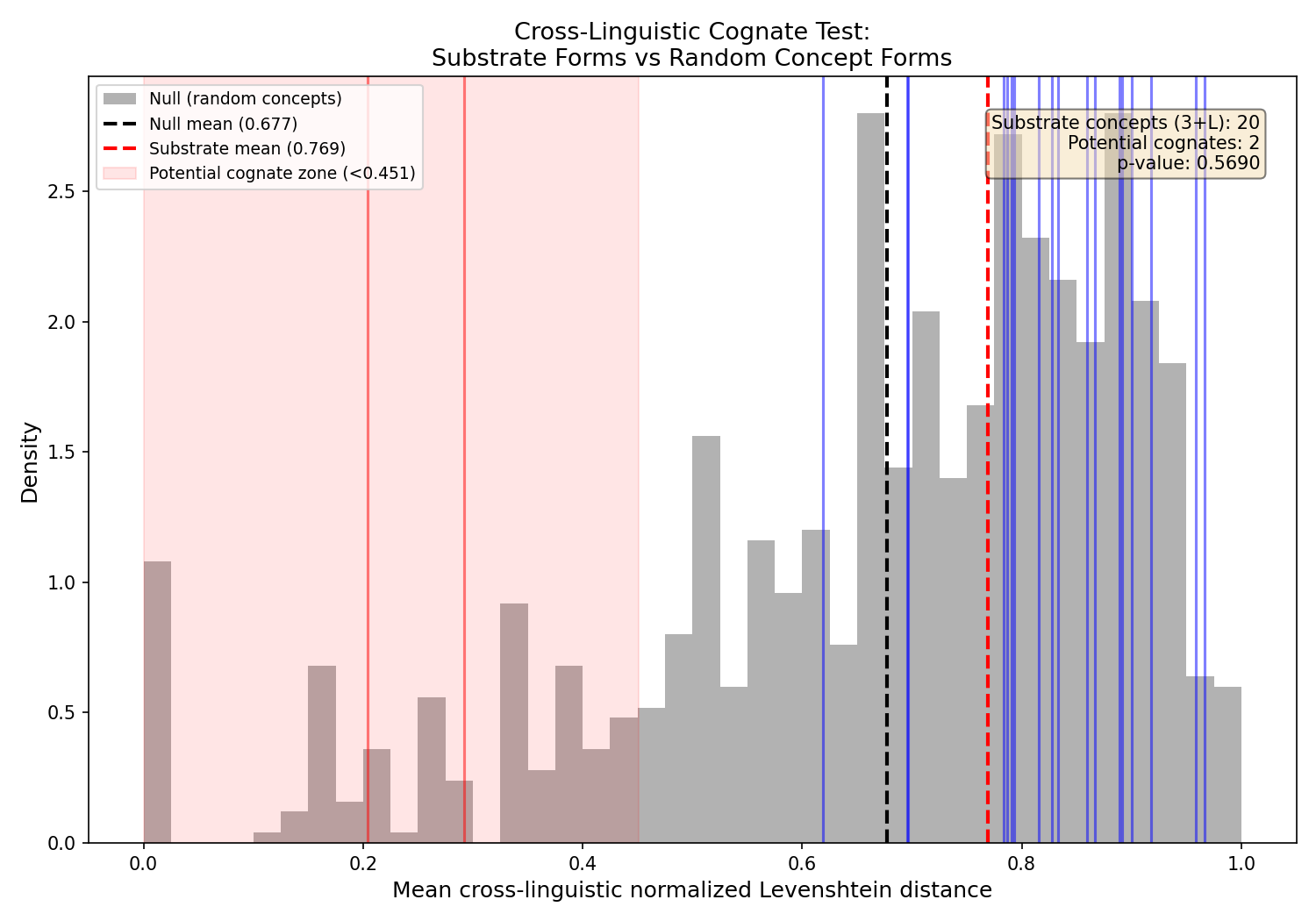}
\caption{Distribution of cross-linguistic pairwise Levenshtein distances for consensus substrate forms (solid line) vs.\ null distribution from 1,000 random concept draws (histogram). Substrate forms are not more similar across languages than expected by chance ($p$ = 0.569).}
\label{fig:crossling}
\end{figure}

For 20 concepts appearing as CS in three or more languages, the mean pairwise Levenshtein distance across languages is 0.769, compared to a null distribution mean of 0.677 ($\pm$ 0.226).
The one-tailed $p$-value is 0.569 (Figure~\ref{fig:crossling}), providing no evidence that substrate forms for the same concept are more similar across languages than random vocabulary.

\subsubsection{Numeral compound false positives}

Two concepts---``Fifty'' and ``Twenty''---show strikingly low cross-linguistic distances (0.204 and 0.292, respectively), initially suggesting shared substrate inheritance.
However, inspection reveals these are transparent Austronesian compounds: \textit{limampulo}/\textit{lima pulu} (``Five'' $\times$ ``Ten''), \textit{ruampulo}/\textit{rua pulu} (``Two'' $\times$ ``Ten''), built from well-established PAn roots *\textit{lima} and *\textit{puluq}.
Both detection methods flagged them because compound numerals have unusual phonological properties (long forms, nasal clusters at morpheme boundaries) that mimic the substrate fingerprint.
This represents a systematic false-positive category that future substrate detection methods should address.
More broadly, morphologically complex Austronesian forms (compounds, derived forms) may mimic the non-mainstream fingerprint through increased length and cluster density at morpheme boundaries.
However, two observations mitigate this concern.
First, non-mainstream candidates show \textit{lower} rates of canonical Austronesian \textit{prefix} patterns (\textit{ma-}, \textit{pa-}, etc.)---though we note that this feature does not capture infixes (\textit{-um-}, \textit{-in-}) or suffixes (\textit{-an}, \textit{-i}) that are also productive in Sulawesi languages, so the morphological profile of non-mainstream forms remains only partially characterized.
Second, removing the length feature entirely from the model produces equivalent performance (CV AUC 0.769, Section~\ref{sec:results_ipa}), demonstrating that the fingerprint does not depend on form length at all---whether measured in characters, syllables, or omitted.

\subsection{Geographic Expansion}
\label{sec:results_expansion}

To test generalization beyond the original six languages, we applied the trained Model~B to 16 additional Indonesian languages from three geographic groups: 8 additional Sulawesi languages, 6 Western Indonesian languages, and 2 Eastern Indonesian languages (Table~\ref{tab:expansion}).

\begin{table}[H]
\centering
\caption{Expansion validation results by geographic group. ``Original'' = the 6 training languages. AUC is computed against each language's own rule-based residual labels.}
\label{tab:expansion}
\small
\begin{tabular}{llrrrrl}
\toprule
Group & $N_\mathrm{langs}$ & Mean Rule\% & Mean ML\% & Mean AUC & Mean $P_\mathrm{sub}$ & Notable outliers \\
\midrule
Original 6     & 6 & 28.0 & 23.1 & 0.890 & 0.326 & --- \\
Sulawesi exp.  & 8 & 49.9 & 62.4 & 0.685 & 0.606 & Bol.Mong. (9.2\%), Gorontalo (84.2\%) \\
W.\ Indonesian & 6 & 28.4 & 35.3 & 0.634 & 0.393 & Acehnese (62.9\%), Sundanese (1.8\%) \\
E.\ Indonesian & 2 & 34.9 & 51.9 & 0.661 & 0.520 & --- \\
\bottomrule
\end{tabular}
\end{table}

\begin{figure}[H]
\centering
\includegraphics[width=0.85\textwidth]{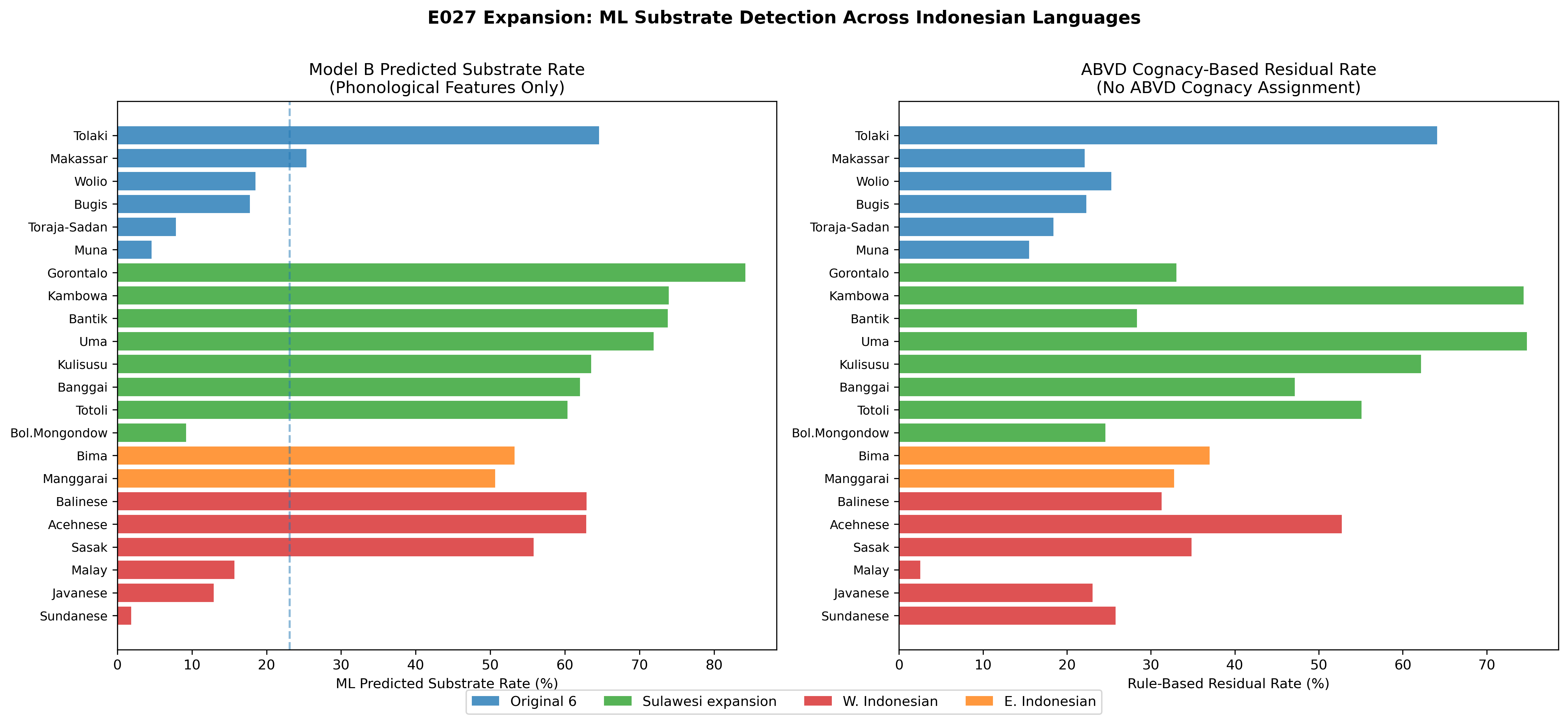}
\caption{Predicted substrate rates (ML) vs.\ rule-based residual rates for 22 languages grouped by geographic region. Languages are ordered by ML substrate rate within each group.}
\label{fig:expansion}
\end{figure}

The mean expansion AUC of 0.663 confirms that Model~B generalizes to unseen languages, though with reduced performance compared to the original training languages (0.890).
The key finding is geographic patterning: Sulawesi expansion languages show significantly higher mean $P_\mathrm{substrate}$ (0.606) than Western Indonesian languages (0.393), a delta of +0.213.

Two outliers merit comment.
Bol.\ Mongondow, despite its Sulawesi location, shows the lowest predicted substrate rate among expansion Sulawesi languages (9.2\%), behaving phonologically like a Western Indonesian language---consistent with strong Gorontalic Austronesian retention.
Acehnese, despite being geographically Western Indonesian, shows a high predicted substrate rate (62.9\%), consistent with its known Chamic heritage and possible Mon-Khmer substrate influence \citep{thurgood1999}.

\section{Discussion}

\subsection{The phonological fingerprint}

The central finding of this study is that a probabilistic phonological profile---which we term a ``fingerprint'' for convenience---distinguishes non-mainstream vocabulary from inherited Austronesian forms in Sulawesi basic lexicon with moderate reliability (AUC = 0.763 after ablation): longer forms, more consonant clusters, higher rates of glottal stops, and fewer canonical Austronesian prefixes.
This profile is detectable by machine learning and generalizes across all six languages in leave-one-out testing (6/6 $\geq$ 0.65 under ablation).

Linguistically, this fingerprint is consistent with what would be expected of substrate vocabulary.
Austronesian basic roots are canonically disyllabic (CVCV or CVCCV), reflecting the Proto-Austronesian root structure \citep{blust2009}.
Forms that deviate from this template---through greater length, consonant clusters, or non-canonical segments like glottal stops---are more likely to originate outside the Austronesian mainstream, whether as pre-Austronesian remnants, areal innovations, or language-specific neologisms.

The prominence of the \textsc{action} semantic domain among high-confidence substrates is noteworthy.
Action verbs may represent a ``vulnerable'' domain for substrate retention: while nouns for basic objects and body parts are readily displaced by incoming Austronesian vocabulary, verbs describing everyday actions may persist because they are more deeply embedded in cultural practice and syntactic frames.
This aligns with observations in creole linguistics that substrate verb semantics tend to be more resistant to superstrate replacement than noun semantics \citep{lefebvre2004, thomason1988}.

\subsection{Parallel innovation, not shared substrate}

The clustering analysis (E029) provides a critical constraint on interpretation.
The 266 consensus substrates do not form coherent word families across languages: silhouette scores are near zero, and the cross-linguistic cognate test is non-significant ($p$ = 0.569).
This provides no support for the hypothesis that these forms derive from a single pre-Austronesian language that was displaced by Austronesian expansion.
Alternative explanations---including deeply diverged substrate(s) beyond the resolving power of Levenshtein distance, or multiple unrelated substrate languages---cannot be excluded, but the parsimonious interpretation is independent innovation.

Instead, the consensus substrates appear to represent parallel independent innovations: each language independently developed non-mainstream vocabulary in similar semantic domains (action verbs, grammatical words), producing forms that share a phonological profile (the fingerprint) without sharing etymological descent.
This is consistent with the ``lexical gap'' interpretation---these are positions in the Swadesh list where each language innovated away from the proto-Austronesian form through language-specific processes (contact, semantic shift, phonological erosion of the original root), rather than inheriting from a common non-Austronesian source.

The numeral compound false positives (``Twenty'', ``Fifty'') illustrate a related methodological lesson: phonological non-conformity is not equivalent to non-Austronesian origin.
Morphologically complex Austronesian forms can mimic the substrate fingerprint, and any computational substrate detection method must account for productive morphology.

\subsection{Tolaki: outlier or genuine divergence?}

Tolaki's dominant contribution to the substrate candidate pool (121/266 CS forms, 45.5\%) demands careful interpretation.
Its low ABVD cognacy coverage (36\%) means that many genuinely Austronesian forms may be misclassified as residuals simply because the ABVD lacks cognate coding for them.
This coverage artifact inflates Tolaki's residual rate and, by extension, its contribution to the ML training set.

However, the LOLO analysis provides partial reassurance: when Tolaki is held out entirely, the model still achieves AUC = 0.806 on Tolaki data---the highest of any language---suggesting that Tolaki's non-mainstream vocabulary has genuinely distinctive phonological properties that the model learns from the other five languages.
The sensitivity analysis ($\Delta$AUC = $-$0.062 without Tolaki) further confirms that the overall result survives Tolaki's removal, though it weakens.

Whether Tolaki's elevated non-mainstream vocabulary reflects genuine pre-Austronesian substrate, greater distance from the South Sulawesi ``mainstream'' represented by Bugis and Makassar, or simply documentation gaps in the ABVD cannot be resolved by computational methods alone and requires targeted fieldwork.

\subsection{Comparison with traditional methods}

Traditional substrate detection in Austronesian linguistics relies on the identification of forms that cannot be reconstructed to any proto-language through regular sound correspondences \citep{blust2009, blust2010}, while recent computational approaches have focused on automated cognate detection \citep{list2012} and large-scale phylogenetic inference from lexical data \citep{jaeger2018}.
This method is powerful but requires deep expertise in each language's historical phonology and scales poorly across large datasets.

Our ML approach is complementary rather than competitive.
It cannot replace the comparative method---it has no access to sound correspondences, morphological decomposition, or diachronic phonological rules.
What it offers is a scalable, reproducible screening tool that can flag phonologically anomalous forms for specialist attention.
The cross-method consensus analysis (Section~\ref{sec:results_consensus}) demonstrates this complementarity: 60.7\% of rule-based residuals are confirmed by ML, while 39.3\% are reclassified as phonologically Austronesian-like (probable false positives), and 41 new candidates are identified that the rule-based method missed.

\subsection{Script-level convergence: the Hanacaraka reduction}
\label{sec:hanacaraka}

An independent line of evidence converges with the computationally detected phonological fingerprint.
The Javanese script Hanacaraka, derived from the Brahmi--Pallava tradition via Kawi, reduces the Sanskrit--Devanagari consonant inventory from 33 to 20 aksara \citep{casparis1975}.
The 13 eliminated consonants fall into exactly the phonological categories that define Austronesian non-conformity in Model~B: aspiration contrasts (8 consonants: \textit{kha}, \textit{gha}, \textit{cha}, \textit{jha}, \textit{tha}, \textit{dha}, \textit{pha}, \textit{bha}), retroflex place distinctions (5: \textit{\d{t}a}, \textit{\d{t}ha}, \textit{\d{d}a}, \textit{\d{d}ha}, \textit{\d{n}a}), and sibilant manner distinctions (three Sanskrit sibilants merged into one).

This script-level reduction is phonologically principled: the categories eliminated from Hanacaraka are precisely those absent from Proto-Austronesian \citep{blust2009}.
PAn reconstructions posit approximately 17 consonant phonemes; Hanacaraka's 20 aksara are a closer match to the PAn inventory than to the 33-consonant Sanskrit source.
The script adaptation reveals the same phonotactic constraints that our ML model detects computationally: when Javanese speakers received an Indic writing system, they stripped away the phonological distinctions that had no correlate in Austronesian phonology.

A further convergence involves the glottal stop.
Javanese has a phonemic glottal stop [\textipa{P}] that is represented in neither Hanacaraka nor Devanagari---it predates the adoption of Indic writing.
Our SHAP analysis identifies glottal stop presence as the fifth most important feature for substrate classification (mean $|$SHAP$|$ = 0.188), with substrate candidates showing significantly higher rates of glottal stops (32.0\%) than consensus Austronesian forms (10.5\%).
That the phoneme most diagnostic of non-mainstream vocabulary is precisely the one absent from both the indigenous and the borrowed script systems suggests it belongs to a deeper phonological layer---consistent with pre-Austronesian or early Austronesian substrate.

While this evidence comes from Javanese rather than the Sulawesi languages in our primary dataset, the Hanacaraka reduction reflects a pan-Western-Malayo-Polynesian phonological constraint.
The convergence between computational substrate detection and historical script adaptation strengthens confidence that the phonological fingerprint identified by Model~B reflects genuine structural properties of the Austronesian phonological system, rather than artifacts of the classification methodology.

\subsection{Limitations}
\label{sec:limitations}

Several limitations constrain the interpretation of these results.

\begin{enumerate}
    \item \textbf{Orthographic, not IPA.} All phonological features were computed from ABVD orthographic forms, which use language-specific orthographic conventions rather than standardized phonemic transcriptions. Our approximate IPA conversion test (Section~\ref{sec:results_ipa}) shows that digraph collapse does not degrade performance ($\Delta$AUC $<$ 0.01), but this conversion is conservative---only 5.5\% of forms were affected. Full IPA conversion with language-specific phonological rules would provide a more thorough validation.

    \item \textbf{Small sample size.} With 1,357 total forms and 438 substrate candidates across six languages, the dataset is small by ML standards. The results should be interpreted as suggestive rather than definitive, and replication on larger vocabularies (beyond Swadesh-210) is needed.

    \item \textbf{Label noise and proxy labels.} The labels index forms that resist attribution to known Austronesian etyma in the ABVD---not forms with confirmed non-Austronesian origin. No ground truth exists for ``substrate'' identity. The Positive-Unlabeled nature of the labeling means that some ``non-mainstream'' candidates are actually Austronesian forms with missing ABVD cognacy data. This label noise biases the classifier toward conservative estimates of the phonological fingerprint's distinctiveness.

    \item \textbf{Cognacy coverage confound.} The top SHAP feature (\texttt{language\_cognacy\_coverage}) is a language-level property correlated with the labeling process. However, the ablation analysis (Section~\ref{sec:results_ablation}) demonstrates that removing this feature slightly \textit{improves} performance (CV AUC +0.003, LOLO AUC +0.007), confirming that the phonological fingerprint is not driven by this confound. The \texttt{language\_id\_encoded} control still contributes modestly: removing both language features reduces CV AUC by 0.033.

    \item \textbf{Geographic scope.} Six languages from a single island cannot represent the full diversity of Austronesian--non-Austronesian contact situations. The expansion analysis (Section~\ref{sec:results_expansion}) provides encouraging generalization, but systematic validation across Melanesia, where Austronesian--Papuan contact is well documented \citep{ross2005, dunn2005}, would be a more rigorous test.

    \item \textbf{No morphological decomposition.} All features operate on surface forms without morphological parsing. Consonant clusters arising from productive morphology (e.g.\ infixation with \textit{-um-}, \textit{-in-}, or suffixation with \textit{-an}, \textit{-i}) are not distinguished from root-internal clusters. Similarly, prefix detection is limited to canonical Austronesian prefixes and does not capture infixes or suffixes. Morphological decomposition of the ABVD forms---using tools such as language-specific analyzers or paradigm-based approaches---would enable root-level feature extraction and a cleaner test of whether the phonological profile is stem-level or morphologically inflated.
\end{enumerate}

\section{Conclusion}

This study demonstrates that machine learning can detect phonological non-conformity in Sulawesi basic vocabulary with moderate reliability (AUC = 0.763 after ablation of confound features), independent of cognacy data.
The resulting phonological fingerprint---longer forms, more consonant clusters, glottal stops, and action-verb semantics---characterizes non-mainstream lexical items that may include pre-Austronesian substrate vocabulary, areal innovations, or language-specific developments.

Cross-method consensus between rule-based subtraction and ML classification identifies 266 high-confidence non-mainstream candidates (Cohen's $\kappa$ = 0.611), while phonological clustering analysis provides the important negative result that these candidates do not form coherent cross-linguistic word families.
The consensus non-mainstream forms are consistent with parallel independent innovations rather than remnants of a single pre-Austronesian language.

Convergent evidence from historical script adaptation strengthens this finding: the Javanese Hanacaraka script, derived from Sanskrit-based Kawi, eliminates exactly the phonological categories (aspiration, retroflexion, sibilant manner) absent from Proto-Austronesian, confirming that the computational model detects genuine Austronesian phonotactic constraints rather than classification artifacts.

Methodologically, the two-model design introduced here---separating circular cognacy-based features from non-circular phonological features---provides a template for computational non-conformity detection that can be extended to other language families and contact situations.
The approach is not a replacement for the comparative method, but a scalable complement that can screen large vocabularies for specialist attention.

Three directions for future work are warranted.
First, although our approximate IPA conversion test (Section~\ref{sec:results_ipa}) confirms robustness, full conversion of ABVD orthographic forms to IPA using tools such as CLTS \citep{anderson2018} would enable phoneme-level feature engineering and eliminate residual orthographic bias.
Second, expansion of the detection pipeline beyond Swadesh-210 to full dictionary data would test whether the phonological fingerprint holds for non-basic vocabulary.
Third, targeted fieldwork on the highest-confidence non-mainstream candidates---particularly Tolaki action verbs---could provide the etymological and areal evidence needed to distinguish genuine pre-Austronesian remnants from independent innovations.

\vspace{1em}
\noindent\textbf{AI Disclosure.}
This research employed an AI-augmented workflow using Claude (Anthropic) for literature synthesis across multiple linguistic databases, statistical scripting (Python, scikit-learn, XGBoost), cross-database comparison (ABVD, CLDF, Proto-Austronesian reconstructions), feature engineering, SHAP analysis, and iterative experimental design.
The AI-augmented approach enabled the researchers to execute six linked experiments (E022--E029, E041) spanning 1,357 lexical forms across 22 languages, with cross-method consensus analysis, phonological clustering, and IPA robustness validation---a throughput that would typically require a larger computational linguistics team.
The research hypotheses (substrate detection via phonological non-conformity, the parallel-innovation alternative), domain interpretation (linguistic significance of the fingerprint, Hanacaraka convergence), and final scholarly judgments were made by the human authors.
All AI-generated code, statistical results, and text were reviewed, validated, and edited by the authors.
The manuscript text was drafted with AI writing assistance and revised by the authors for accuracy and scholarly voice.

\vspace{1em}
\noindent\textbf{Data availability.} All data are from the publicly available Austronesian Basic Vocabulary Database \citep{greenhill2008}. Analysis scripts and feature matrices are available at GitHub: \url{https://github.com/neimasilk/volcarch-repo} (MIT licence).

\bibliography{references}

\end{document}